# The Modeling of SDL Aiming at Knowledge Acquisition in Automatic Driving

**Zecang Gu**

(Apollo Japan Co., Ltd. CTO gu@apollo-japan.ne.jp)

**Yin Liang**

(China University of GeoScience, liangyin@cug.edu.cn)

**Zhaoxi Zhang**

(Shijiazhuang Tiedao University, zhaoxi_zhang@163.com)

## Abstract

In this paper we proposed an ultimate theory to solve the multi-target control problem through its introduction to the machine learning framework in automatic driving, which explored the implementation of excellent drivers' knowledge acquisition. Nowadays there exist some core problems that have not been fully realized by the researchers in automatic driving, such as the optimal way to control the multi-target objective functions of energy saving, safe driving, headway distance control and comfort driving, as well as the resolvability of the networks that automatic driving relied on and the high-performance chips like GPU on the complex driving environments. According to these problems, we developed a new theory to map multi-target objective functions in different spaces into the same one and thus introduced a machine learning framework of SDL(Super Deep Learning) for optimal multi-target control based on knowledge acquisition. We will present in this paper the optimal multi-target control by combining the fuzzy relationship of each multi-target objective function and the implementation of excellent drivers' knowledge acquired by machine learning. Theoretically, the impact of this method will exceed that of the fuzzy control method used in automatic train.

## Introduction

The automatic driving is one of the most outstanding and promising application areas of artificial intelligence, but there is few concern on the method of constructing the knowledge acquisition model of excellent drivers and the optimal multi-target control, which are essential for the automatic driving.

The most representative control method in automatic driving is using networks. The information, delivered by the Autonomous Highway Vehicle(AHV) about its status and predetermined heading, such as ID, time, location, orientation, speed, length, width, predetermined route, etc. to the servers through networks, will compose the spatial decisive data to help the network servers directly control the AHV[1][2]. However, the automatic driving technology of vehicles is far more complex than that of trains. For example, the right moment and the appropriate distance for two driving cars to keep away from each other are more fuzzy degree than object data, while in practice the object data are necessary for the strict machine control. Furthermore, the objective functions of automatic driving, which not only refer to the pilot process but also involve more issues about safe driving, headway distance control, comfort driving etc., are separately defined in the different spaces so that it is dramatically difficult to directly control them only based on networks.

Another representative automatic driving method is building the control system with high-performance chips in order to process the world-scale data online[3]. Nevertheless, since the automatic driving system have to face complex road and driving conditions and also try to solve the optimization problems of safe driving, headway distance control, comfort driving simultaneously, the control system equipped with high-performance chips will be incapable of optimizing the multi-target functions based on fuzzy degree.

## 1. Some barriers automatic driving needs to break through

In this section, we will have a discussion on some kernel problems in automatic driving about human-machine judgment conflict, human-machine sensory integration, and human-machine authority transfer etc., so as to explore the new theoretical methods and build the efficient models for the optimization of the multi-target functions[4].

**Human–Machine judgment conflict:**

This February, in the Google's automatic driving experiment, when the automatic vehicle met a barrier in the front and tried to turn right for the avoidance, a truck came from the right-back and then a heavy accident happened because the truck's driver thought that the Google's vehicle should stop in front of the barrier.

**Human–Machine Sensory Integration:**

According to the investigation of a company, when meeting a automatic vehicle, there are 41% people think the best way is to stay away as far as they can, while others suggest that keeping a fixed or a little near distance is better. And even more there are a few persons with curiosity to catch up with the front vehicle. This is a knotty problem in automatic control to integrate the sensory deviations between machines and human beings and guide the vehicles in the closest way to people's driving.

**Human–Machine authority transfer:**

Senses can not transfer with authorities between human beings and machines. For instance, it is easy to miss an opportunity when machines chose different plans from human beings in emergency when authorities need to be transferred.

**Trolley problem:**

The famous Trolley Problem is described as how to reduce the number of sacrifices to the lowest in emergency. It is related to not only intricate ethical issues, but also technical difficulties so that nobody has proposed a valuable solution to the Trolley problem of automatic driving.

**Fuzzy control:**

People are aware that there is a determined control point which makes the automatic driving different. Therefore, it is impossible for the traditional methods to solve the problem with multi-target functions, such as safe driving, headway distance control and comfort driving, while the speed of the automatic vehicle is changing in the complicated road. The only solution to the multi-target control is to rely on the experience of excellent drivers, which is what the machine learning model is skilled in.

## 2. The model of Super Deep Learning(SDL)

While the main characteristic of the last AI climax is the application of expert system, the machine learning model will be the center of a vortex in this climax of AI and should be used to solve probabilistic problems. Based on this core, the large-scale popularization of machine learning will be carried out so that its application effect will surpass that of any other technology on society in the past. No matter how high the evaluation is, it will not be excessive.

The Figure 1 is the schematic diagram of best classification for probability space. As shown in the Figure 1, the main problem, which AI need to solve nowadays, is to get the probability distribution of the object function through machine learning and then decide which probability distribution is closest to a point in the space, that is solving the problem of best classification of probability space in the Euclidean space.

In the Figure 1, assuming there are two centers, w and v, which belong to the different probability spaces composed by the probability distribution W and V in the Euclidean space E, there is a point r and the major task of machine learning is to figure out which is closest to r, w or v. This proposition is a typical pattern recognition problem and can be easily solved by working out the rigorous distance formula between w and v. As the kernel problem in machine learning, the distance between probability distributions is getting more and more attention from many AI researchers[5][6][7].

Based on the theory proposed by the Russian mathematician Andrey Kolmogorov: "the probability space is measurable space with the measure of 1", the lemma 1 can be proposed as: "there is only one probability distribution in the probability space, and the distance in probability space is a way to solve the distance between the Euclidean space and the probability space". Data used by AI are located in one of probability spaces in the Euclidean space, so the classification criteria in AI should be dependent on the distance leaping over the Euclidean space to the probability space.

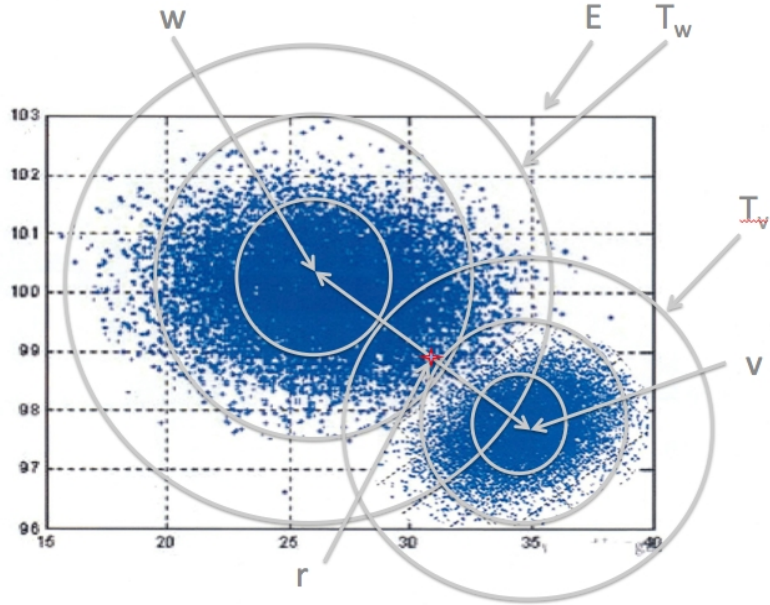

**Figure 1. the best classification for probability spaces**

Based on the lemma 1 mentioned above, the lemma 2 can be introduced as: the distance, whose probability distribution is 1 in the probability space, is zero.

The formula about the distance leaping over the Euclidean space to the probability space can be developed according to the lemma 2. Assuming there two points, $w_j$ (j=1, 2, …, n) belonging to the vector W as the center of the probability distribution $T_w$, and $v_j$ (j=1, 2, …, n) belonging to the vector V as the center of the probability distribution $T_v$, the distance between $v_j$ and $w_j$ leaping over the Euclidean space and the probability space is as follows:

$$G(W,\ V) = \left\{\sum_{j=1}^{n}(w_j - v_j)^2\right\}^{\frac{1}{2}} \quad (1)$$

$$(w_j - v_j) = \begin{cases} 0 & |w_j - v_j| \leq \Delta_j^{(vj)} \\ |w_j - v_j| - \Delta_j^{(vj)} & |w_j - v_j| > \Delta_j^{(vj)} \end{cases} \quad (2)$$

$$\Delta_j^{(vj)} = \sum_{i=1}^{m_{ij}^{(vj)}} D_{ij}^{(vj)} P_{ij}^{(vj)} \quad (3)$$

After entering the probability space $T_w$ from the Euclidean space E, the direction is from $w_j$ to $v_j$ and $\Delta_j^{(vj)}$ is the distance error, where $D_{ij}$ and $P_{ij}$ are values of the Euclidean distance $D_{ij}^{(vj)}$ and the probability distance $P_{ij}^{(vj)}$ in the distance segment of No.i (i=1,2….,$m_{ij}^{(vj)}$). This formula means the probability distribution of each segment is independent of the distance error. This accords with the fact that the distance scale in the probability space has no symmetry[7]. Imitating the above method, it is obvious that we can calculate the distances from the point r in the Euclidean space to the point v in the probability space $T_v$ and the point w in the probability space $T_w$ separately.

The Figure 2 is the schematic diagram of two space-map methods of solving complex problems. As the Figure 2(a) showed, mapping a complex problem to n spaces can be called the actinoid mapping[8]~[12]. For example, in order to extract and recognize the deep information around a car, the environment image can be mapped into many mapping images by means of geometric and physical models. The Figure 2 (b) is the schematic diagram of mapping several spaces into one and it is called the astriction mapping[8]~[12]. For example, in the problem of multi-objective control, it is hard to find a optimal control point for many object functions in different spaces. With the astriction mapping, it is possible to map the multi-objective functions into a membership fuzzy space so as to get the optimal control point in the unified form.

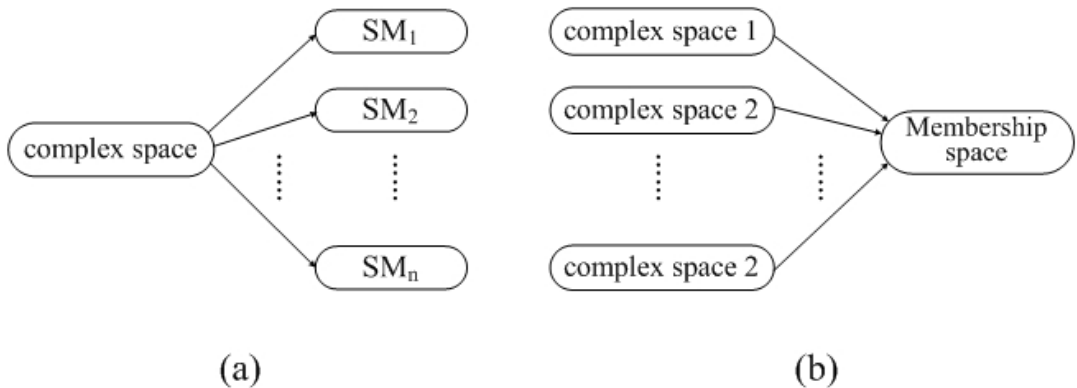

Figure 2 two mapping methods of complex space n

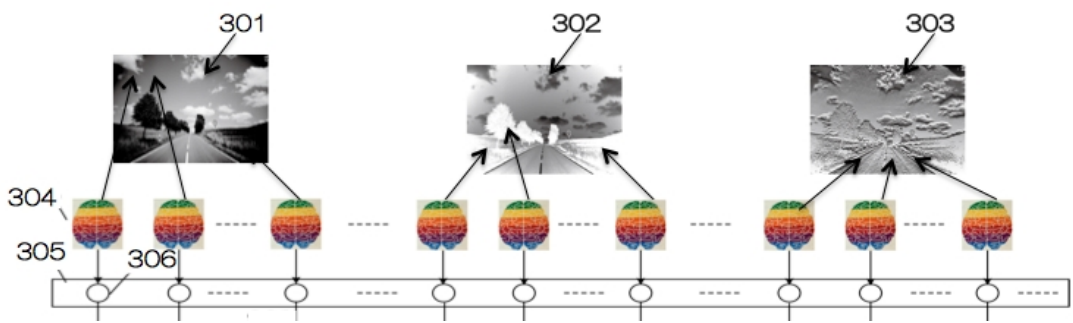

Figure 3 the actinoid mapping suitable for environment images of automatic vehicles

The Figure 3 is the schematic diagram of the actinoid mapping suitable for environment images of automatic vehicles. There are two key steps that will influence the accuracy of image recognition: information extraction and pattern recognition.

As the Figure 3 showed, 301 is the raw image around a car, while 302 and 303 are the mapping images of raw images separately. The deep information in the environmental images of cars can be extracted by applying the geometric and physical mapping models. 304 is obtained by the eigenvalues extracted from some area of the mapping images through unsupervised leaning. 305 is a perception layer in neural network and 306 is a node in this perception layer. In pattern recognition the quality of information extraction is more important than the machine-learning classification algorithm. Based on the space mapping theory, the deep information in images surrounding cars and the high-order maximum probability of eigenvalues, which is able to represent the regional features and transcend the values derived from traditional statistics, can be extracted directly by mapping one image to many ones with geometric or physical models.

As shown in the Figure 3, in order to extract more information from automobile's environmental images, it is necessary to make use of various space mapping for the environmental images and obtain more region segments to distill the features for each mapping image. In the model of figure 3.1, the nodes of perception layers are able to be added infinitely so that it is possible to recognize the automobile images in the arbitrary depth, which will not introduce the extra complexity(only around $O(n^2)$) and aggravate the difficulties of image recognition surrounding automobiles. Therefore, it is prospective that this model is more reasonable and capable of exceeding the traditional deep learning that increases the recognition effects by adding the number of hidden layer nodes.

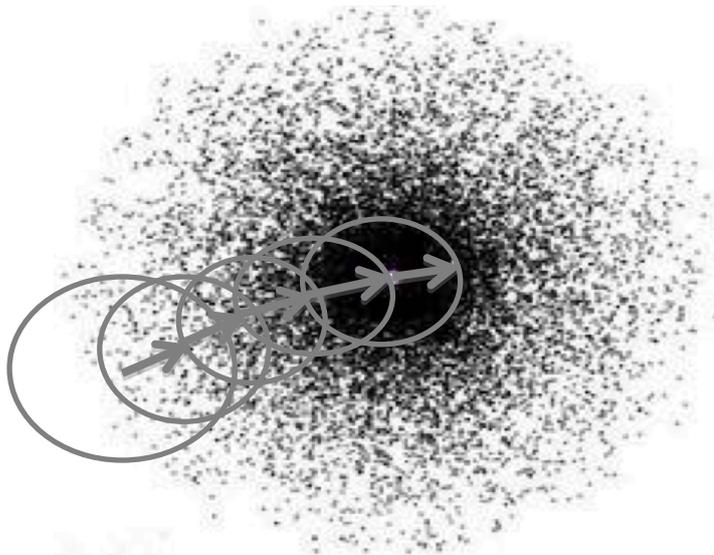

**Figure 4 the schematic diagram of machine learning model**

As showed in Figure 4, assuming there is a set of probability distributions belonging to a probability space in the Euclidean space, the number of elements in the set is $\zeta(g_f \in G, (f=1,2,\ldots,\zeta))$ and there must exist a characteristic value $A(G)$ in $g_f$ ($f=1, 2, \ldots, \zeta$). Since the probability space is a measure space, a probability scale $M[G, A(G)]$ must exist to satisfy the conditions below, and it can be taken as the reference to make iterations, which will make G migrate to the direction of the maximum probability.

$$A^{(n)} = A(G^{(n)}) \quad (4)$$

$$M^{(n)} = M[G^{(n)}, A(G^{(n)})] \quad (5)$$

$$G^{(n)} = G\{[A(G^{(n-1)}), M[G^{(n-1)}, A(G^{(n-1)})]]\} \quad (6)$$

$A(G^{(n)})$ can be set as the characteristic value with the maximal probability, and $M[G^{(n)}, A(G^{(n)})]$ can be the maximal probability scale whose center is $A(G^{(n)})$. The unsupervised machine-learning model showed in the Figure 4 is based on the theory of Self-organizing with a probability scale. It is because of the introduction of maximal probability scale $M[G^{(n)}, A(G^{(n)})]$ and the self-organizing calculation method taking $M[G^{(n)}, A(G^{(n)})]$ as a reference that the model of Self-organizing based on a probability scale can get a breakthrough.

The maximal probability scale $M[G^{(n)}, A(G^{(n)})]$ is the probability scale containing the probability attributes of any one of normal distribution, multivariable normal distribution, logarithm normal distribution, exponential distribution, t distribution, F distribution, $X^2$ distribution, binomial distribution, minus binomial distribution, multinomial distribution, Poisson distribution, Erlang distribution, hypergeometric distribution, geometric distribution, traffic distribution, Weibull distribution, triangular distribution, Beta distribution, Gamma distribution.

The maximal probability scale M is also the scale about statistical properties or distances of probability spaces, including Mahalanobis distance, functions based on Gaussian Processes, Wasserstein Distance scale, KL(Kullback-Leibler) Distance scale, P (Pearson) Distance scale, Probability Measure of Fuzzy Event, etc.

The maximal probability scale M can also be introduced as the Euclidean Distance scale, such as Manhattan Distance scale, Chebyshev Distance scale, Minkowski Distance scale, Cosine scale according to non-probability space.

While acting as any one of Jaccard-similarity Coefficient scale, such as Hamming Distance scale or Information Entropy, it is another extended example for the maximal probability scale M to be used in the Bayesian Analysis method to do self-organizing between the prior probability and the posterior probability.

The features of unsupervised machine-learning model showed in the Figure 4 are the little effect of the initial position, the migration to the maximal direction by itself, the spanning solution of high-order maximal probability over the traditional statistics, the directly obtainable probability distribution information of objective functions, the distributed machine-learning system, etc.

## 3. The implementation of knowledge acquisition with SDL in automatic driving

In this section, we will make a concrete introduction to the implementation of excellent drivers' knowledge acquisition with SDL.

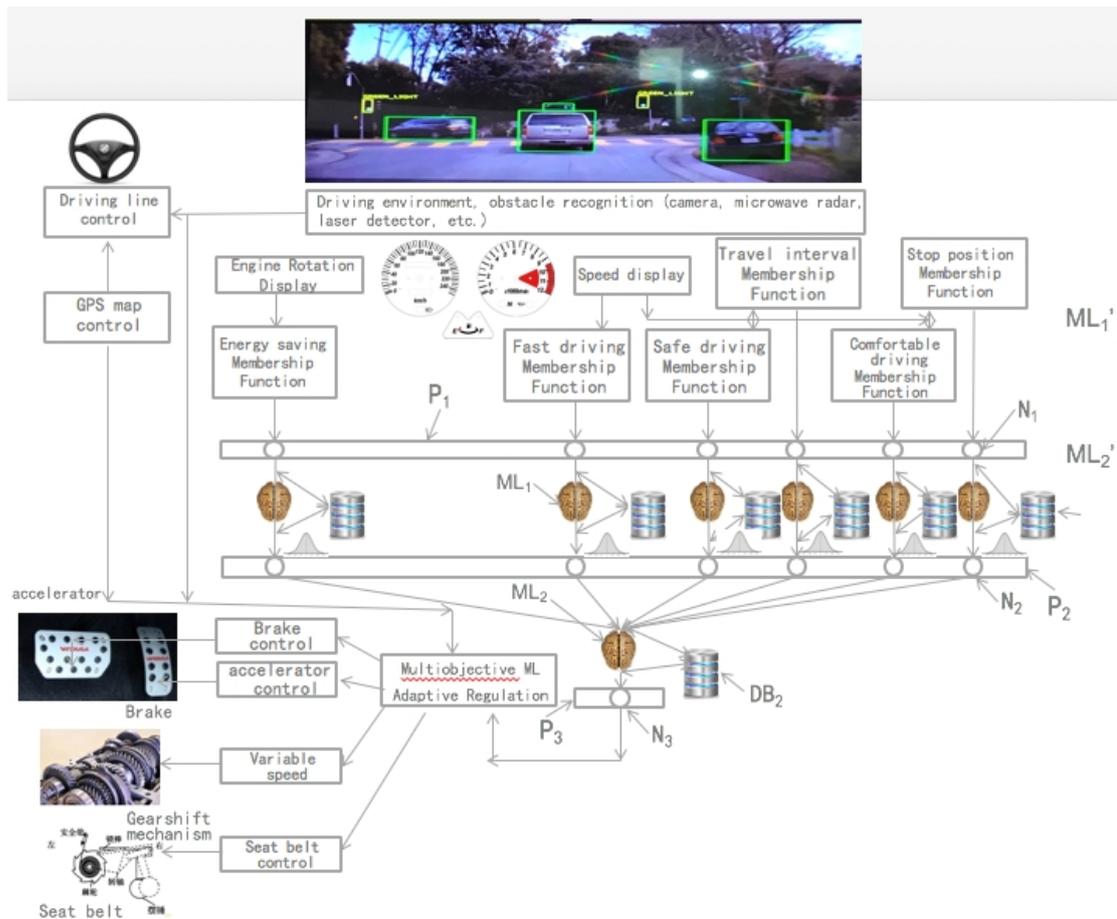

**Figure 5 the machine-learning model of optimal multi-target control**

As the Figure 5 showed, the problems of energy saving, safe driving, headway distance control, comfort driving etc. are situated in the different spaces. There are three issues needing to be solved in the construction of machine-learning model in the optimal multi-target control :

1. mapping multi-target functions from the different spaces to the same one;
2. using machine learning to implement the knowledge acquisition of excellent drivers to imitate their driving skills;
3. regulating multi-target functions online automatically and rapidly by means of machine learning.

We will mainly focus on the issue 2 in this paper, while the issue 1 and the issue 3 will be emphatically discussed in another paper.

Based on the machine learning through excellent drivers' driving for many times, the values reflected by energy saving, safe driving, headway distance, comfort driving, quickly driving and stopping distance will form a high-order maximal probability distribution function via the corresponding unsupervised machine learning $ML_1$, which puts the data got from the knowledge acquisition into the database $DB_1$.

There are many states for automatic vehicles, such as:

State 1: the car runs forward, and there are no barriers and other cars around it;

State 2: the car runs forward, and there is a car on the left behind;

State 3: the car runs forward, and there is a car on the left front;

State 4: the car runs forward, and there is a car on the left;

State 5: the car runs forward, and there is a car on the right behind;

State 6: the car runs forward, and there is a car on the right front;

State 7: the car runs forward, and there is a car on the right;

State 8: the car runs forward, and there is a car in the front;

State 9: the car runs forward, and there is a car on the back;

… …

State n: …

Each state is corresponding to a different result acquired from the multi-target control machine learning, so the driving state should be determined before the unsupervised machine learning and every state should be corresponding to this state. In other words, the main purpose of multi-target optimal control machine learning in automatic driving is to achieve a probability distribution from the driving data of excellent drivers by many times machine learning based on a set of fuzzy values of multi-target optimal control in each driving state.

The figure 6 is a schematic diagram of knowledge acquisition of excellent drivers implemented by machine learning. As showed in the Figure 6, in order to implement the knowledge acquisition, each probability distribution processed by machine learning, from state 1 to state n, correspond to each result of the multi-target optimal control machine learning. All the states, from (a) to (f), represent energy saving, safe driving, headway distance, comfort driving, quickly driving and stopping distance separately. Actually, each of them reflect the corresponding probability distribution of each state. The greater the degree of probability dispersion is, the less the strict requirement of the objective function of this state is, and vice versa. The states of automatic driving are more than 9 and the method mentioned here can cover all the states, from 1 to n, and so is the objective functions.

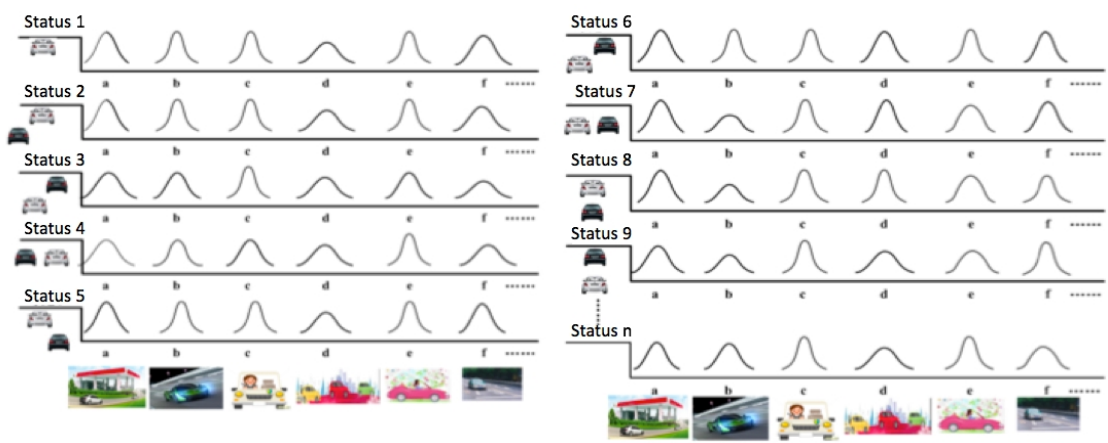

**Figure 6 the knowledge acquisition of excellent driver implemented by machine learning**

In the normal process of automatic driving, a state can produce a set of objective function values, which will be sent to each nodes (N1) of the perception layer (P1) in the Figure 5. The results of every objective function, that is the probability distributions obtained by processing the objective functions represented in the Figure 6 from (a) to (f)through the machine learning ($ML_1$) in the Figure 5, will be used to calculate the distance from a node in the Euclidean space to the center of probability distribution in the probability space based on the Figure 1, the formula 2 and 3, which leaps over the Euclidean space to the probability space. At last, the final result will be sent to every node (N2) of the neural layer (P2) in the Figure 5.

Based on the probability scale acquired by processing the final result with every node (N2) in the neural layer (P2) in the Figure 5 through the self-organizing machine learning, it is possible for us to make a decision to adjust the object function with the maximal probability scale. Meanwhile it is necessary to introduce a threshold $\zeta$ adjusted by hand to estimate whether the distance of the machine learning result will exceed it or not. If the number of objective functions whose distances got from the machine learning (ML2) exceed $\zeta$ is odd, adjustments will be applied to these functions. If it is even, the objective function with the maximal distance will be adjusted and it is also acceptable to adjust an odd number of objective functions whose distances exceed the threshold at the same time by sending the useful information of them to the nodes (N3) in the neural layer (P3).

In order to fully control the braking mechanism, the accelerator mechanism, the speed control mechanism and the safety belt, one of control methods, including linear-system control, nonlinear-system control, optimal control, stochastic control, adaptive control, fuzzy control, qualitative control, predictive control, real-time expert control, etc., should be introduced into the adjustment information in the nodes (N3) of the neural layer (P3).

However, since the force of braking mechanism is correlated to the speed of cars, the different speeds have to match the different forces of the braking mechanism. It is a waste of time to adjust the braking mechanism repeatedly in the self-organize way of the traditional methods, which makes the performance of emergency response inefficient. In order to make it meet the qualification, the parameters of self-organize control can be used as the input data in machine learning and then put them into the control system.

So is the accelerator mechanism. Every time using the data gotten by the machine learning based on the parameters that accelerator mechanism needs for the self-organize control, the speed of control can be accelerated.

All the operations above are implemented in the center of self-organize control in the base of the multi-target machine learning. Each adjustment above needed to return to the machine learning (ML1, ML2) to judge whether it is valid or not. To get the same effect as excellent drivers can drive, it is necessary to carry out the circle of learning and adjusting for many times. The center of self-organize control based on the multi-target machine learning is also able to control the steering wheel according to the result of lane line recognition and route marked in GPS.

The environment recognition is also implemented by the dispersed machine learning. The Figure 3 showed the process of mapping environment images to several mapping images: SM1, SM2…by using physical or geometrical models, but the accordance with practical needs should also be considered. For instance, in cognitive recognition we need to map the lane line in environment images at first. And then the mapping images will be divided into many regions, while every region corresponds to an unsupervised learning ML1. At last, the high-order maximal probability eigenvalues extracted from the corresponding region will be assigned to the node N1 in the perception layer P1.

The unsupervised learning ML2 between the node N1 and the node N2 in the perception layer P1 and P2 must have the capacity of obtaining the probability distribution from the same region of the same image by many times learning and storing it in the database DB1.

The unsupervised learning ML3 between the node N2 and the node N3 in the perception layer P2 and P3 has the function of recognizing images and is able to use environment images collected by cameras and data about the probability distribution from the DB1 to calculate the distance leaping over the Euclidean space to the probability space. Generally the image recognition should accord to practical needs, just like lane lines, burst images, etc.

## Conclusion

In this paper we proposed an implementation of knowledge acquisition based on machine learning in automatic driving, which can implement multi-target optimal control theory in automatic driving according to excellent drivers' skill. The characteristics of this method is to carry out the dispersed machine learning and implement the optimal control in each step with SDL model. To extract information of environment images, a space mapping theory of images is developed with the capacity of mining the deep information of images and recognizing the harsh environment of automobiles accurately. In term of the integral control of automatic driving, a theory of implementing the multi-target optimal control based on skills of excellent drivers has been built, whose effect may exceed that of the fuzzy inference method used in automatic trains. Theories proposed in this paper still need to be experimented and confirmed by the concrete automatic driving platform.